\documentclass{article}
\usepackage{spconf,amsmath,graphicx}

\usepackage[utf8]{inputenc} 
\usepackage[T1]{fontenc}    
\usepackage{hyperref}       
\usepackage{url}            
\usepackage{booktabs}       
\usepackage{amsfonts}       
\usepackage{nicefrac}       
\usepackage{microtype}      
\usepackage{multirow,rotating}

\usepackage{xspace}

\usepackage{listings}
\usepackage{color}

\usepackage[scaled]{beramono}
\definecolor{pythonfile}{rgb}{0,0.6,0.8}
\definecolor{codegreen}{rgb}{0,0.6,0}
\definecolor{codegray}{rgb}{0.5,0.5,0.5}
\definecolor{codepurple}{rgb}{0.58,0,0.82}
\definecolor{backcolour}{rgb}{0.95,0.95,0.92}

\lstdefinestyle{mystyle}{
    backgroundcolor=\color{backcolour},   
    commentstyle=\color{codegreen},
    keywordstyle=\color{magenta},
    numberstyle=\tiny\color{codegray},
    stringstyle=\color{codepurple},
    basicstyle=\ttfamily\scriptsize,
    breakatwhitespace=false,         
    breaklines=true,                 
    captionpos=b,                    
    keepspaces=true,                 
    numbers=left,                    
    numbersep=5pt,                  
    showspaces=false,                
    showstringspaces=false,
    showtabs=false,                  
    tabsize=2
}
\newcommand{\nop}[1]{}

\title{Microsoft Dialogue Challenge: \\ Building End-to-End Task-Completion Dialogue Systems}
\name{Xiujun Li\quad Yu Wang\quad Siqi Sun\quad Sarah Panda\quad Jingjing Liu\quad Jianfeng Gao}
\address{Microsoft, Redmond, WA, 98052, USA}
\begin{document}
%
\maketitle
\begin{abstract}
This proposal introduces a Dialogue Challenge for building end-to-end task-completion dialogue systems, with the goal of encouraging the dialogue research community to collaborate and benchmark on standard datasets and unified experimental environment. In this special session, we will release human-annotated conversational data in three domains (movie-ticket booking, restaurant reservation, and taxi booking), as well as an experiment platform with built-in simulators in each domain, for training and evaluation purposes. The final submitted systems will be evaluated both in simulated setting and by human judges.
\end{abstract}

\section{Introduction}
\label{sec:introduction}


There are many virtual assistants commercially available today, such as Apple's Siri, Google's Home, Microsoft's Cortana, and Amazon's Echo. With a well-designed dialogue system as an intelligent assistant, people can accomplish tasks easily via natural language interactions. 

In the research community, dialogue system has been well studied for many decades. Recent advance in deep learning has also inspired the exploration of neural dialogue systems. However, it still remains a big challenge to build and evaluate multi-turn task-completion systems in a universal setting. 

On one hand, conversational data for dialogue research has been scarce, due to challenges in human data collection and privacy issues. Without standard public datasets, it has been difficult for any group to build universal dialogue models that could encourage follow-up studies to benchmark upon. On the other hand, labeled datasets that are available now, while useful for evaluating partial components of a dialogue system (such as natural language understanding, dialogue state tracking), fail at end-to-end system evaluation. As a thorough evaluation of a dialogue system requires a large number of users to interact with the system at real time. 

A well-adopted alternative approach is the employment of user simulators. The idiosyncratic strength and weakness of simulators for dialogue systems has been a long-standing research topic. User simulators can provide an interactive environment for evaluating dialogue system, which is more attainable and less costly than human evaluation. The use of simulators can also foster interest and encourage research effort in exploring reinforcement learning for dialogue management. 

However, the progress of dialogue research via reinforcement learning is not as fast as we have expected, largely due to the lack of a common evaluation framework, on which different research groups can jointly develop new technologies and improve their systems. In addition, the dependency on simulators often limits the scope of functionality of the implemented dialogue systems, due to the inevitable discrepancy between real users and artificial simulators. Over the past few years, we have achieved some initial success in this area. This proposal aims to further develop and mature this work and release a universal experimentation and evaluation framework by working together with research teams in the community.

In this proposal, we present a new Dialogue Challenge on ``End-to-End Task-Completion Dialogue System''. This differs from previous dialogue tracks, most of which have focused on component-level evaluation. In this dialogue challenge, we will release a carefully-labeled conversational dataset in multiple domains. This data can be used by participants to develop all the modules required to build task-completion dialogue systems. We will also release an experimentation platform with built-in simulators. Each domain will have its own well-trained simulator for experimentation purpose.

In the rest of the proposal, Section~\ref{sec:platform_overview} will provide more details about the proposed experimentation platform. Section~\ref{sec:task_des} will describe the specific tasks defined in the challenge, as well as the corresponding datasets that will be released. And Section~\ref{sec:eval} will describe the final evaluation of submitted systems. 

\begin{figure*}[ht]
\centering
\includegraphics[width=\linewidth]{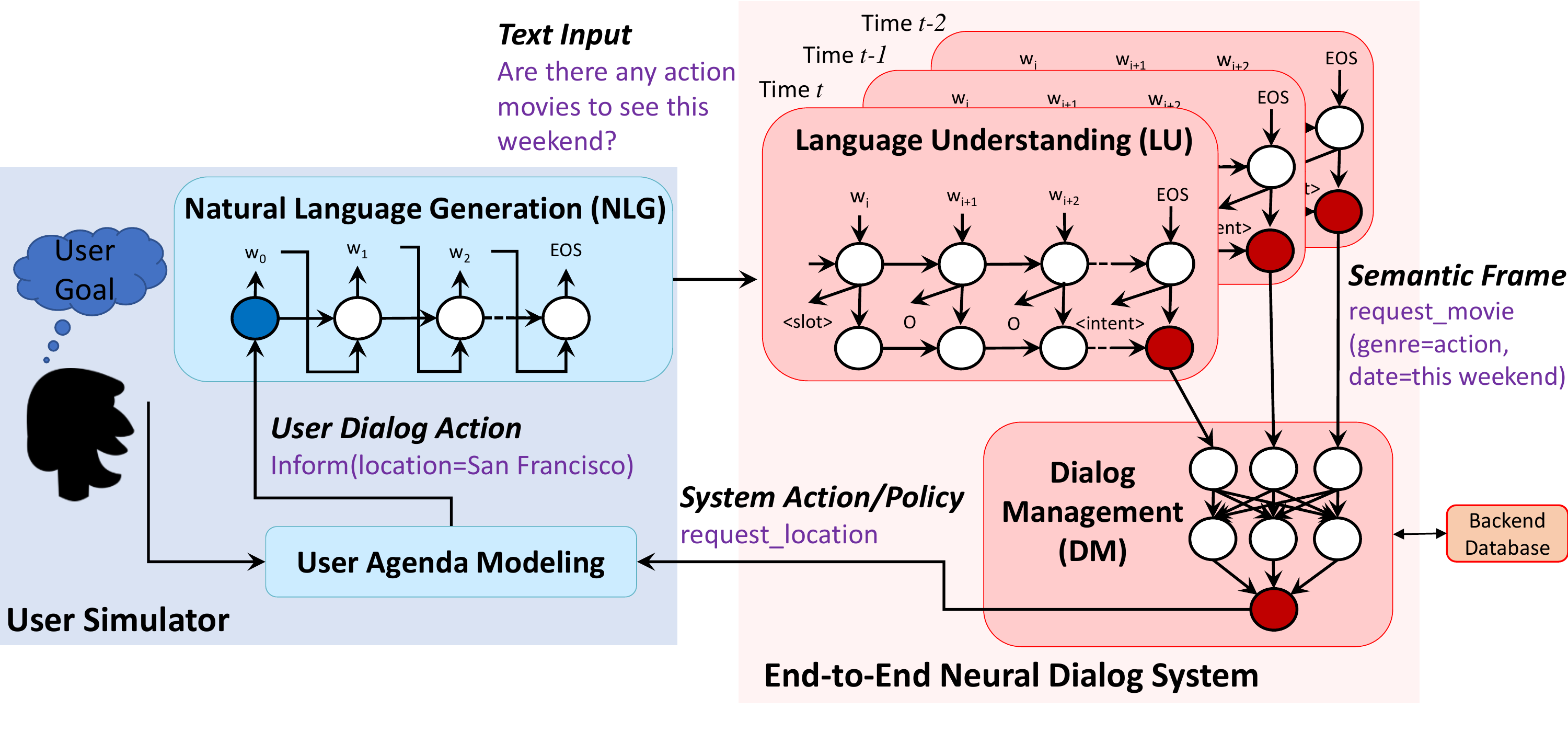}
\vspace{-8mm}
\caption{Illustration of an end-to-end task-completion dialogue system.}
\vspace{-2mm}
\label{fig:end2end}
\end{figure*}

\section{Platform Overview}
\label{sec:platform_overview}
The proposed experimentation platform is illustrated in Figure~\ref{fig:end2end}~\cite{li2017end}. It consists of a user simulator (on the left) that mimics a human user and a dialogue system (on the right). In the user simulator, an agenda-based user modeling component~\cite{schatzmann2009hidden} works at the dialog-act level, and controls the conversation exchange conditioned on the generated user goal, to ensure that the user behaves in a consistent, goal-oriented manner. An NLU (Natural Language Understanding) module will process user's natural language input into a semantic frame. An NLG (Natural Language Generation) module is used to generate natural language sentences responding to the user's dialogue actions. 

Although Figure~\ref{fig:end2end} presents a neural dialogue system as an example, participants are free, and encouraged, to plug in any NLU/NLG modules, as long as their systems can complete a predefined task via multi-turn conversations with the user. In every turn of a conversation, the system needs to understand natural language input generated by the user or the simulator, track dialogue states during the conversation, interact with a task-specific dataset (described in Section~\ref{sec:task_des}), and generate an action (i.e., system response). The system action could be presented either as semantic frames (known as \emph{dialog-acts} in simulation evaluation), or as natural language utterances generated by an NLG module.

\section{Task Description}
\label{sec:task_des}

In this dialogue challenge, we will release well-annotated datasets for three task-completion domains\footnote{All the datasets and code will be available at \url{https://github.com/xiul-msr/e2e_dialog_challenge}.}: movie-ticket booking, restaurant reservation, and taxi ordering. Table~\ref{tab:stat_tasks} shows the statistics of the three datasets. We will use movie ticket booking as an example to explain the specific task of building dialogue system in each domain.

\begin{table}[h]
\small
\begin{tabular}{|c|c|c|c|c|}
\hline
Task & \#Intents & \#Slots & \#Dialogs \\
\hline
Movie-Ticket Booking & 11 & 29 & 2890 \\
\hline
Restaurant Reservation & 11 & 30 & 4103 \\
\hline
Taxi Ordering & 11 & 19 & 3094 \\
\hline
\end{tabular}
\centering
\vspace{1mm}
\caption{The Statistics of Three Tasks.}
\label{tab:stat_tasks}
\end{table}

\subsection{Movie-Ticket booking task}
In this task, the goal is to build a dialogue system to help users find information about movies and book movie tickets. Throughout the course of the conversation, the agent gathers information about the user’s requests, and in the end books the movie tickets or provides information about the movie in question. At the end of the conversation, the dialogue environment assesses a binary outcome (success or failure), based on (1) whether a movie ticket is booked and (2) whether the movie satisfies the user’s constraints.

\subsubsection{Dataset}
The data that will be released for this task was collected via Amazon Mechanical Turk. The annotation schema contains $11$ intents (e.g., inform, request, confirm\_question, confirm\_answer, etc.), and $29$ slots (e.g., moviename, starttime, theater, numberofpeople). Most of the slots are \textit{informational} slots, which can be used to constrain the search. Others are \textit{request} slots, with which users can request information from the agent. 
The final dataset to release will consist of 2890 dialogue sessions, with approximately $7.5$ turns per session on average. Table \ref{tab:data_example} shows an example of annotated human-human dialogue in the movie-ticket booking task. And Table \ref{tab:tcp_sample} shows one success and one failure dialogue example, generated by a rule-based agent and an RL agent interacting with user simulator, respectively. 

\subsubsection{User-Goal Set}
The user goals that will be released alongside with the labeled data, are extracted from labeled dialogues by two methods. The first one extracts all the slots (known and unknown) from the first user turn (excluding the greeting turn) in each session, under the assumption that the first turn usually contains the main request from the user. The second method aggregates all the slots that appear in the user turns into one user goal. These user goals are then stored into a user-goal database for the simulator to draw from. When triggering a dialogue, the user simulator randomly samples one user goal from this database. Figure 2 shows one example user goal for the movie-ticket booking task. 

\begin{figure}[htb]
\begin{lstlisting}[language=XML]
New episode, user goal:
{
  "request_slots": {
    "ticket": "UNK"
  },
  "inform_slots": {
    "city": "seattle",
    "numberofpeople": "2",
    "theater": "amc pacific place 11 theater",
    "starttime": "9:00 pm",
    "date": "tomorrow",
    "moviename": "deadpool"
  }
}
\end{lstlisting}
\label{fig:user_goal_example}
\vspace{-2mm}
\caption{An example of a user goal: the user wants to buy 2 tickets of Deadpool at 9:00 PM tomorrow at amc pacific place 11 theater{,} Seattle.}
\end{figure}

\subsubsection{Knowledge Base}
The knowledge base to be released for this task was built from labeled dialogues. The data entries are organized in JSON files, which include several KBs of size 1000 records or above. Examples of these data records are as follows:

\begin{figure}[htb]
\begin{lstlisting}[language=XML]
#movie 1
'city': 'hamilton', 'theater': 'manville 12 plex', 'zip': '08835', 'critic_rating': 'good', 
'date': 'tomorrow', 'state': 'NJ', 'starttime': '10:30am', 'genre': 'comedy', 'moviename': 'zootopia'

#movie 2
'city': 'seattle', 'theater': 'regal meridian 16', 'zip': '98101', 'theater_chain': 'regal meridian',
'state': 'WA', 'starttime': '6:30pm', 'date': 'tonight', 'moviename': 'zootopia'
\end{lstlisting}
\label{fig:mv_kb_exmaples}
\vspace{-2mm}
\caption{Data records in Movie Knowledge Base.}
\end{figure}

\subsubsection{User Simulator}
For the experimentation platform, we will also release a user simulator~\cite{li2016user} for this task. The user simulator can support two formats of input: 
\begin{enumerate}
\item \emph{Frame-level semantics}: A dialog act form (e.g., \textsf{request(moviename; genre=action; date=this weekend)}) that can be used for debug purpose.
\item \emph{Natural language}: Natural language text. To use this format, each participate needs to build their own NLU component to convert natural language input into frame-level semantics.
\end{enumerate}

\section{Evaluation}
\label{sec:eval}
To evaluate the quality of the submitted systems, we will conduct both simulation evaluation and human evaluation.

\subsection{Simulation Evaluation}
Three metrics will be used to measure the quality of the systems: \{\emph{success rate
, average turns, average reward}\}. Success rate is sometimes known as \emph{task completion rate} -- the fraction of dialogues that ended successfully. Average turns is the average length of the dialogue. Average reward is the average reward received during the conversation. There is a strong correlation among the three metrics: generally speaking, a good policy should have a high success rate, high average reward and low average turns. Here, we choose \emph{success rate} as our major evaluation metric. 

\subsection{Human Evaluation}
We will also conduct human evaluation for the competition. We will ask human judges to interact with the final systems submitted by participants. Besides the measurements aforementioned, each user will also give a rating on a scale of 1 to 5 based on the naturalness, coherence, and task-completion capability of the system, at the end of each dialogue session.



\begin{table*}[htbp!]
\small
\centering
\caption{Annotated human-human dialog examples for the movie-ticket booking task and the restaurant reservation task. Left column shows natural language. Right column shows annotated dialog act form. (\underline{usr}: User, \underline{agt}: Agent)}
\label{tab:data_example}
\begin{tabular}{l|l}
\hline

\multicolumn{1}{c|}{\textbf{Natural language}} & \multicolumn{1}{c}{\textbf{Dialog Act}} \\
\hline
\multicolumn{2}{c}{\textbf{Movie-Ticket Booking Task}} \\
\hline

\underline{usr}: Find me a good action movie this weekend.  & \underline{usr}: request(moviename;genre=action;date=this weekend)  \\[3pt]
\underline{agt}: \begin{tabular}[c]{@{}l@{}} London Has Fallen is currently the number 1 \\ action movie in America. \end{tabular} & \underline{agt}: \begin{tabular}[c]{@{}l@{}} inform(moviename=london has fallen; other=\\number 1; genre=action) \end{tabular} \\ [8pt]

\underline{usr}: Oh that sounds terrific.  & \underline{usr}: confirm\_answer()  \\ [3pt]
\underline{agt}: \begin{tabular}[c]{@{}l@{}} Would you like to purchase tickets to this movie? \\ I would need to know what city you are in. \end{tabular} & \underline{agt}: request(city) \\ [8pt]
\underline{usr}: Seattle & \underline{usr}: inform(city=seattle) \\ [3pt]
\underline{agt}: \begin{tabular}[c]{@{}l@{}} Which theater in Seattle should I search for \\ tickets? \end{tabular} & \underline{agt}: request(theater;city=seattle)  \\ [8pt]
\underline{usr}: Regency or AMC please. & \underline{usr}:inform(theater\_chain={amc\#regency})   \\ [3pt]
\underline{agt}: \begin{tabular}[c]{@{}l@{}} Around what time do you want to go, \\ and on which day? \end{tabular}  &  \underline{agt}: request(date) \\ [8pt]
\underline{usr}: 9:30 pm any day this week. &  \underline{usr}: inform(starttime=9:30 pm;date=this week)  \\  [3pt]
\underline{agt}: \begin{tabular}[c]{@{}l@{}} London Has Fallen is showing at 9:45pm on \\ Wednesday at AMC Southcenter 16, is that \\ showing acceptable for you? \end{tabular}  &  \underline{agt}: \begin{tabular}[c]{@{}l@{}} inform(moviename=london has fallen; \\ starttime= 9:45pm; date=wednesday; theater=\\ amc southcenter 16)  \end{tabular}  \\ [13pt]
\underline{usr}: yes that would be great.	&    \underline{usr}: confirm\_answer()     \\ [3pt]
\underline{agt}: Excellent, how many tickets would you like?  &  \underline{agt}: request(numberofpeople)   \\ [3pt]
\underline{usr}: I want 2 tickets please! &  \underline{usr}: inform(numberofpeople=2) \\ [3pt]
\hline

\multicolumn{2}{c}{\textbf{Restaurant Reservation}} \\
\hline
\underline{usr}: \begin{tabular}[c]{@{}l@{}}Im looking for a martini bar in Indianapolis. \end{tabular}  & \underline{usr}: \begin{tabular}[c]{@{}l@{}} request(restaurantname;food=martini bar;\\city=Indianapolis) \end{tabular} \\ [3pt]
\underline{agt}: \begin{tabular}[c]{@{}l@{}} Here is the restaurant I found: High Velocity. \\ Do you want to book? \end{tabular} & \underline{agt}: \begin{tabular}[c]{@{}l@{}} request(reservation;restaurantname=High Velocity) \end{tabular} \\ [8pt]
\underline{usr}: \begin{tabular}[c]{@{}l@{}} YES \end{tabular}  & \underline{usr}: \begin{tabular}[c]{@{}l@{}} confirm\_answer() \end{tabular} \\ [3pt]
\underline{agt}: \begin{tabular}[c]{@{}l@{}} at what date would you like to go? \end{tabular} & \underline{agt}: request(date) \\ [3pt]
\underline{usr}: \begin{tabular}[c]{@{}l@{}} saturday night \end{tabular} & \underline{usr}: inform(date=Saturday night) \\ [3pt]
\underline{agt}: \begin{tabular}[c]{@{}l@{}} at what time would you like to go? \end{tabular} & \underline{agt}: request(starttime) \\ [3pt]
\underline{usr}: \begin{tabular}[c]{@{}l@{}} 8pm \end{tabular} & \underline{usr}: inform(starttime=8pm) \\ [3pt]
\underline{agt}: \begin{tabular}[c]{@{}l@{}} how many people are going? \end{tabular} & \underline{agt}: request(numberofpeople) \\ [3pt]
\underline{usr}: \begin{tabular}[c]{@{}l@{}} 4 \end{tabular} & \underline{usr}: inform(numberofpeople=4) \\ [3pt]
\underline{agt}: \begin{tabular}[c]{@{}l@{}} Your reservation at High Velocity for 02/27/2016 \\ 08:00PM for 4 people under Joe Does has been \\ confirmed. \end{tabular}  & \underline{agt}: \begin{tabular}[c]{@{}l@{}} inform(taskcomplete;restaurantname=High Velocity;\\date=02/27/2016;starttime=08:00pm;numberofpeople=4;\\personfullname=Joe Does)  \end{tabular} \\ [3pt]
\hline

\end{tabular}
\end{table*}

\begin{table*}[htbp!]
\small
\centering
\caption{Two sample dialogs generated by a rule-based agent and an RL agent with user simulator. Left column shows scenarios where both the rule-based and RL-based agents succeed. Right column shows scenarios where the rule-based agent fails, while the RL agent succeeds.(\underline{usr}: User, \underline{agt}: Agent)}
\label{tab:tcp_sample}
\begin{tabular}{l|l}
\hline
\begin{tabular}[c]{@{}l@{}}
New episode, user goal:\\
\{\\
\-\hspace{3mm} "request\_slots": \{	\\
\-\hspace{6mm}    "ticket": "UNK"	\\
\-\hspace{3mm}  \},	\\
\-\hspace{3mm}  "diaact": "request",	\\
\-\hspace{3mm}  "inform\_slots": \{	\\
\-\hspace{6mm}   "city": "seattle",	\\
\-\hspace{6mm}    "numberofpeople": "2",	\\
\-\hspace{6mm}    "theater": "regal meridian 16",	\\
\-\hspace{6mm}    "starttime": "9:25 pm",	\\
\-\hspace{6mm}    "date": "tomorrow",	\\
\-\hspace{6mm}    "moviename": "zoolander 2"	\\
\-\hspace{3mm}  \}\\
\}\\
\end{tabular}  & \begin{tabular}[c]{@{}l@{}}
New episode, user goal:\\
\{\\
\-\hspace{3mm} "request\_slots": \{	\\
\-\hspace{6mm}    "ticket": "UNK"	\\
\-\hspace{6mm}    "theater": "UNK"	\\
\-\hspace{6mm}    "starttime": "UNK"	\\
\-\hspace{3mm}  \},	\\
\-\hspace{3mm}  "diaact": "request",	\\
\-\hspace{3mm}  "inform\_slots": \{	\\
\-\hspace{6mm}    "numberofpeople": "3",	\\
\-\hspace{6mm}    "date": "tomorrow",	\\
\-\hspace{6mm}    "moviename": "10 cloverfield lane"	\\
\-\hspace{3mm}  \}\\
\}\\
\-\hspace{3mm}
\end{tabular}                \\
\hline
\multicolumn{1}{c}{\textbf{Rule-based agent}} & \multicolumn{1}{c}{\textbf{Rule-based Agent}} \\
\hline

\underline{usr}: Can I get some tickets for zoolander 2?  & \underline{usr}: What is the start time for 10 cloverfield lane?  \\
\underline{agt}: What movie are you interested in? & \underline{agt}: What movie are you interested in?  \\
\underline{usr}: I want to watch zoolander 2.  & \underline{usr}: I want to watch at 10 cloverfield lane.  \\
\underline{agt}: What time would you like to see it? & \underline{agt}: What time would you like to see it? \\
\underline{usr}: I want to watch at 9:25 pm. & \underline{usr}: What time is it playing tomorrow? I need 3 tickets. \\
\underline{agt}: What city you would like? & \underline{agt}: What city you would like?  \\
\underline{usr}: I want to watch at seattle. & \underline{usr}: I do not care.   \\
\underline{agt}: What date would you like to watch it? &  \underline{agt}: What date would you like to watch it? \\
\underline{usr}: I want to set it up tomorrow. &  \underline{usr}: I want to set it up tomorrow.  \\   
\underline{agt}: Which theater would you like?   &    \underline{agt}: Which theater would you like?    \\
\underline{usr}: I want to watch at regal meridian 16.	&    \underline{usr}: Which theater is available?      \\
\underline{agt}: How many tickets do you need?  &  \underline{agt}: How many tickets do you need?   \\
\underline{usr}: I want 2 tickets please! &  \underline{usr}: I want 3 tickets please! \\ 
\begin{tabular}[c]{@{}l@{}}\underline{agt}: Great - I was able to purchase 2 tickets for \\ you to see zoolander 2 tomorrow at regal meridian 16 \\ theater in seattle at 9:25 pm. \end{tabular}  &  \begin{tabular}[c]{@{}l@{}}\underline{agt}: Okay, I have booked 3 tickets for 10 cloverfield \\ lane tomorrow. \end{tabular}     \\
\underline{usr}: Thank you. & \underline{usr}: Thank you. \\ 
\underline{agt}: Thank you. & \underline{agt}: Thank you. \\ 
\hline
\multicolumn{1}{c|}{\textbf{Success}} & \begin{tabular}[c]{@{}l@{}} \textbf{Failure:} Agent does not answer all the questions \\ (starttime, theater) from user.\end{tabular}
\\ \hline

\multicolumn{1}{c}{\textbf{RL agent}} & \multicolumn{1}{c}{\textbf{RL Agent}} \\
\hline

\underline{usr}: Can I get tickets for zoolander 2 tomorrow?  & \begin{tabular}[c]{@{}l@{}}\underline{usr}: Which theater can I book 3 tickets for \\ 10 cloverfield lane?  \end{tabular} \\
\underline{agt}: Which city would you like?    & \underline{agt}: What time would you like to see it?  \\
\underline{usr}: I want to watch at seattle.    & \begin{tabular}[c]{@{}l@{}} \underline{usr}: Which theater and start time are available \\ tomorrow?  \end{tabular}\\
\underline{agt}: How many tickets do you need?  & \underline{agt}: 11:45am is available.     \\
\underline{usr}: I want 2 tickets please!      & \underline{usr}: Which theater is available? \\
\underline{agt}: 9:25 pm is available.       & \underline{agt}: regal la live stadium 14 is available.  \\
\underline{usr}: I want to watch at regal meridian 16.   & \underline{usr}: Could you help me to book the tickets?  \\
\begin{tabular}[c]{@{}l@{}}\underline{agt}: Great - I was able to purchase 2 tickets for \\ you to see zoolander 2 tomorrow at regal meridian 16 \\ theater in seattle at 9:25 pm. \end{tabular}   &   \begin{tabular}[c]{@{}l@{}} \underline{agt}: Okay - I was able to book 3 tickets for you to \\ see 10 cloverfield lane at regal la live stadium 14 at \\ 11:45am tomorrow. \end{tabular}      \\
\underline{usr}: Thank you.    &      \underline{usr}: Thank you  \\   
\underline{agt}: Thank you.   &      \underline{agt}: Thank you     
\\ \hline
\multicolumn{1}{c|}{\textbf{Success}} & \multicolumn{1}{c}{\textbf{Success}}
\\ \hline

\end{tabular}
\end{table*}


\bibliographystyle{IEEEbib}
\bibliography{refs}

\begin{thebibliography}{1}

\bibitem{li2017end}
Xuijun Li, Yun-Nung Chen, Lihong Li, and Jianfeng Gao,
\newblock ``End-to-end task-completion neural dialogue systems,''
\newblock {\em arXiv preprint arXiv:1703.01008}, 2017.

\bibitem{schatzmann2009hidden}
Jost Schatzmann and Steve Young,
\newblock ``The hidden agenda user simulation model,''
\newblock {\em IEEE transactions on audio, speech, and language processing},
  vol. 17, no. 4, pp. 733--747, 2009.

\bibitem{li2016user}
Xiujun Li, Zachary~C Lipton, Bhuwan Dhingra, Lihong Li, Jianfeng Gao, and
  Yun-Nung Chen,
\newblock ``A user simulator for task-completion dialogues,''
\newblock {\em arXiv preprint arXiv:1612.05688}, 2016.

\end{thebibliography}

\end{document}